# Beyond Two Bytes per Letter: Tokenization Overhead in Cyrillic AI Systems

*Ivan Dobrovolskyi, Independent Researcher*

## Abstract

Modern multilingual tokenizers often fragment Ukrainian and other underrepresented Cyrillic-script languages more heavily than English, creating disparities in cost and context capacity. We quantify this overhead across nine production tokenizers and five languages with standardized Cyrillic and Latin representations, covering 8.37 million word forms. On a corpus benchmark, Ukrainian shows 68–121% token overhead on modern tokenizers and 220% on the older cl100k, measured through full-text fertility on the BrUK and Brown corpora. Overhead is negatively associated with Cyrillic vocabulary allocation in the subset with independently verified English baselines, although the association is not statistically significant (Spearman $\rho = -0.536$, $p = 0.215$, $n = 7$). We evaluate two mitigation strategies. LLMLingua-2 reduces Ukrainian input length by 47–49% on an e-commerce RAG benchmark of 1,536 products and 145 queries, with no compression-induced value losses among 80 retrievable cases. A balanced byte-level BPE tokenizer trained with a 200K vocabulary cap, converging at 158,184 actual entries, reduces the held-out UK/EN ratio from 2.22× to 1.30×. Romanization increases Ukrainian token counts by 2–19% on most tokenizers. Across the five languages, tokenization efficiency favors the script more prevalent in web data. These findings indicate that training data allocation contributes to Cyrillic tokenization overhead and that mitigation is possible at both inference and tokenizer-design stages.

## 1 Introduction

Byte-Pair Encoding (BPE) tokenizers learn subword merge rules from training corpora that are predominantly English and Latin-script (Petrov et al., 2023). Languages written in Cyrillic script face a compounding disadvantage: standard Cyrillic code points (U+0400–U+04FF) require two bytes in UTF-8 encoding, whereas ASCII Latin letters require one. (Latin letters with diacritics also require two bytes, but these are infrequent in English.) The resulting Cyrillic byte sequences appear less frequently in predominantly English training data, producing fewer and shorter merge rules. The result is higher token counts per word — a direct multiplier on API cost, inference latency, and effective context window.

Prior work has documented this disparity at the corpus level. Petrov et al. (2023) showed that low-resource languages require up to 15× more tokens than English. Ahia et al. (2023) quantified the economic implications. Maksymenko and Turuta (2025) measured Ukrainian-specific overhead at 1.7–3.3× across foundational models. Churchill and Skiena (2026) proposed post-hoc vocabulary additions for languages where individual characters are missing from tokenizer vocabularies.

We extend this line of work in four directions. First, we provide a cross-linguistic analysis across five languages for which standardized Cyrillic and Latin representations were available (Ukrainian, Serbian, Kazakh, Azerbaijani, Moldovan) totaling 8.37 million word forms, showing that the overhead consistently tracks the dominant web script for each language. Second, we train balanced BPE tokenizers that reduce the UK/EN ratio from 2.22× to 1.30× under a 200K vocabulary cap, with the best balanced tokenizer converging at 158,184 actual entries ($p < 0.0001$), providing experimental evidence that training-data allocation can substantially reduce the overhead and that UTF-8 byte length alone does not determine tokenization efficiency. Third, we evaluate prompt compression as an inference-time mitigation. Fourth, we validate on a production-like benchmark of 1,536 real e-commerce products with deterministic accuracy evaluation, moving beyond corpus-level measurement to an applied setting. The study provides controlled evidence that training-data composition substantially affects Ukrainian tokenization efficiency and demonstrates complementary mitigation strategies at inference and tokenizer-training stages.

Ukrainian serves as the primary case study because it has active deployment of AI assistants in commercial applications and an ongoing public debate about potential script reform (romanization). Our findings generalize to other underrepresented Cyrillic-script languages and inform tokenizer design for multilingual systems.

## 2 Related Work

### 2.1 Tokenization Fairness

Petrov et al. (2023) established that BPE tokenizers introduce systematic unfairness, with fertility rates varying by an order of magnitude across languages. Ahia et al. (2023) connected fertility to economic cost, showing that non-English users pay more per semantic unit. Maksymenko and Turuta (2025) provided the first Ukrainian-specific measurement, benchmarking nine tokenizers on the BrUK corpus. We use their Ukrainian-specific benchmark as the methodological starting point for Study 1 and extend it to newer 2026 tokenizers, while adding cross-script evaluation, prompt compression, and balanced tokenizer training as separate mitigation experiments.

### 2.2 Vocabulary Modification

Churchill and Skiena (2026) proposed adding per-character tokens for scripts where individual Unicode characters are absent from the vocabulary, demonstrating improved hidden-state similarity on Llama 3.2 1B for 12 low-resource languages. Other work has explored broader vocabulary adaptation strategies. Downey et al. (2023) compared methods for specializing multilingual subword vocabularies and initializing embeddings for new languages, while Csaki et al. (2024) showed that target-language vocabulary expansion can reduce tokenizer fertility and improve inference efficiency during language adaptation. Ukrainian differs from the languages targeted by Churchill and Skiena (2026): Cyrillic characters exist in all modern tokenizer vocabularies, but the available subword merges remain insufficient. Our balanced tokenizer experiment addresses this problem by retraining a byte-level BPE tokenizer with proportional English–Ukrainian language representation rather than modifying the vocabulary of an already pretrained language model.

### 2.3 Prompt Compression

LLMLingua (Jiang et al., 2023) introduced perplexity-based token pruning for prompt compression. LLMLingua-2 (Pan et al., 2024) reformulated compression as token classification using XLM-RoBERTa, achieving 2–5× compression with minimal quality loss on English benchmarks. Selective Context (Li et al., 2023) uses self-information for token selection, and RECOMP (Xu et al., 2024) trains extractive and abstractive compressors. We chose LLMLingua-2 because its multilingual backbone (XLM-RoBERTa) makes it directly applicable to Ukrainian without language-specific fine-tuning. To our knowledge, this is the first evaluation of prompt compression on a Cyrillic-script language.

### 2.4 Information-Theoretic Context

Lavreniuk et al. (2026) measured Ukrainian character-level entropy at 1.20 bits per character (76.4% redundancy), comparable to English at approximately 1.0–1.3 bits (Shannon, 1951). These estimates provide an information-theoretic context for interpreting the results, although they were obtained from different corpora and procedures and are not directly comparable.

## 3 Data and Methods

For a tokenizer t and corpus c, fertility is defined as $F(t,c) = N_tokens(t,c) / N_words(c)$. Corpus-level Ukrainian overhead is defined as $R(t) = F(t,BrUK) / F(t,Brown)$. For the Ukrainian romanization experiment, we report $\Delta_rom = (N_tokens(romanized) - N_tokens(Cyrillic)) / N_tokens(Cyrillic) \times 100$, so positive values mean that romanization increases token count relative to native Cyrillic. For the cross-linguistic script comparison, we report $\Delta_cyr = (N_tokens(Cyrillic) - N_tokens(Latin)) / N_tokens(Latin) \times 100$, so positive values mean that Cyrillic is more expensive than Latin. These two quantities answer different questions and are therefore reported separately.

### 3.1 Tokenizer Benchmark (Study 1)

We tokenized the BrUK corpus — 1.34 million Ukrainian words across 805 texts in balanced genres (fiction, journalism, academic, official) — with nine commercial tokenizer encodings: o200k, Llama 4, Gemma 4, Qwen 3/3.5, Mistral Small, Grok, Claude 4.5/4.6, and cl100k. For tokenizers with independently available English measurements, the English baseline uses the Brown corpus (1.23 million words using the same word-counting method as BrUK; genre-matched) accessed via NLTK. For each tokenizer, we recorded the tokenizer source, implementation route, and access date. Public tokenizers were evaluated locally where available; Claude 4.5/4.6 was measured through API token counting only, so English baseline and vocabulary statistics are not reported for Claude. Vocabulary statistics are therefore restricted to tokenizers with locally inspectable vocabularies. UK/EN ratios and correlation analyses were computed only for complete cases, defined as tokenizers with both an independently reproducible Brown-corpus English baseline and locally inspectable Cyrillic vocabulary statistics under comparable measurement conditions. Tokenizers missing one of these components are reported descriptively but excluded from ratio-based analyses.

### 3.2 Cross-Linguistic Dictionary Benchmark (Study 2)

We compiled word-form dictionaries for five languages for which standardized Cyrillic and Latin representations were available: Ukrainian

(lang-uk, 2.78M word forms), Serbian (2.05M), Moldovan (1.56M), Kazakh (1.17M), and Azerbaijani (0.80M) — 8.37 million word forms total. Each dictionary was tokenized in both its native Cyrillic script and a romanized Latin variant using official transliteration standards. For Ukrainian, we tested three systems: KMU 55 (passport), DSTU 9112:2021 System A (diacritics), and DSTU 9112:2021 System B (digraphs). Processing used a Rust-based parallel profiler with the Rayon library (780 seconds on Apple Silicon for all languages and tokenizers). Web-dominant script status was assigned based on the predominant contemporary online written form of each language; because this is used as a categorical interpretive variable rather than a directly measured corpus statistic, we report it descriptively.

### 3.3 E-Commerce RAG Benchmark (Study 3)

We retrieved structured product data for 1,536 items across 10 categories from a major Ukrainian e-commerce platform's public API. Each record includes title, price, brand, seller, availability, technical specifications, review count, and rating. We generated 145 deterministic queries across seven types (price lookup, top-3 by rating, budget filter, product comparison, brand filter, cheapest, and most expensive) with expected answers derived from the API data. Retrieval uses paraphrase-multilingual-MiniLM-L12-v2. Compression uses LLMLingua-2 (microsoft/llmlingua-2-xlm-roberta-large-meetingbank). Token counting covers six tokenizers: o200k, Llama 4, Gemma 4, Qwen 3.5, Mistral, and Grok.

### 3.4 Balanced Tokenizer Experiment

We trained byte-level BPE tokenizers using the HuggingFace tokenizers library on size-matched corpora (BrUK and Brown, 821K words each after matching). We used an 80/20 train/test split (seed 42) and report all results on held-out test data. We trained balanced 50/50 English–Ukrainian tokenizers at four vocabulary sizes: 32K, 50K, 100K, and 200K. At the 100K vocabulary size, we additionally conducted a composition sweep over 11 Ukrainian shares ranging from 0% to 100% in ten-percentage-point increments.

### 3.5 Accuracy Evaluation

Factual accuracy was evaluated deterministically on all 145 queries by checking whether expected values (prices, ratings, brand names) appear in the retrieved and compressed context. We used query-aware retrieval depth (top_k=3 for comparisons, 5 for price lookups, 10–15 for category-wide searches). Additionally, end-to-end validation was performed on all 80 retrievable queries by generating responses with Gemini 2.5 Flash from both raw and compressed contexts and checking whether the expected value appeared in the model's answer.

## 4 Results

### 4.1 Tokenization Overhead

Table 1 reports Ukrainian tokens per word across nine commercial tokenizers and UK/EN ratios for the seven tokenizers with independently verified English baselines. Among tokenizers with verified English baselines, the overhead ranges from 1.68× (Llama 4) to 3.20× (cl100k). The six modern tokenizers with verified English baselines cluster between 1.68× and 2.21×, while cl100k reaches 3.20×. Qwen 3 shows high Ukrainian fertility (2.82 tokens/word) and a small Cyrillic vocabulary. We report these values descriptively, but omit the UK/EN ratio because a matching Brown-corpus English baseline was not available under the same complete-case protocol.

| Tokenizer | UK tok/w | EN tok/w | UK/EN | Cyr vocab | CPT |
|---|---|---|---|---|---|
| Llama 4 Scout | 1.75 | 1.04 | 1.68× | 1,439 | 3.5 |
| Gemma 4 | 1.84 | 1.03 | 1.78× | 13,395 | 4.8 |
| o200k | 1.96 | 1.03 | 1.90× | 14,208 | 4.4 |
| Mistral Small | 2.01 | 1.06 | 1.91× | 748 | 3.2 |
| Qwen 3.5 | 2.01 | 1.05 | 1.92× | 1,194 | 3.5 |
| Grok (xAI) | 2.27 | 1.03 | 2.21× | 3,068 | 3.7 |
| Claude 4.5/4.6 | 2.42 | — | — | — | — |
| Qwen 3 | 2.82 | — | — | 496 | 2.8 |
| cl100k | 3.33 | 1.04 | 3.20× | 729 | 3.1 |

*Table 1: Tokens per word on the BrUK corpus (1.34M words) and the Brown corpus (1.23M words, paper method). CPT = mean characters per Cyrillic vocabulary entry. Claude: UK measured via API; EN and vocabulary statistics omitted because no public tokenizer is available for independent verification. Qwen 3: Ukrainian fertility and vocabulary statistics are reported descriptively; the UK/EN ratio is omitted because a matching Brown-corpus English baseline was not available under the same complete-case protocol.*

The overhead is associated with Cyrillic vocabulary allocation in the complete-case subset: tokenizers with more Cyrillic merge rules tend to produce fewer tokens per word, although the association is not statistically significant (Spearman ρ = −0.536, p = 0.215, n = 7; Claude and Qwen 3 were excluded because one or more required components for the complete-case analysis were unavailable under comparable measurement conditions). The relationship is not purely linear — Llama 4 achieves the lowest overhead (1.68×) with only 1,439 Cyrillic tokens, indicating that vocabulary count alone does not determine tokenization efficiency. Vocabulary size should be viewed as a coarse proxy for script allocation; merge quality, training corpus composition, merge ordering, and tokenizer architecture may also influence fertility. The mechanism behind Llama 4's efficiency remains an open question.

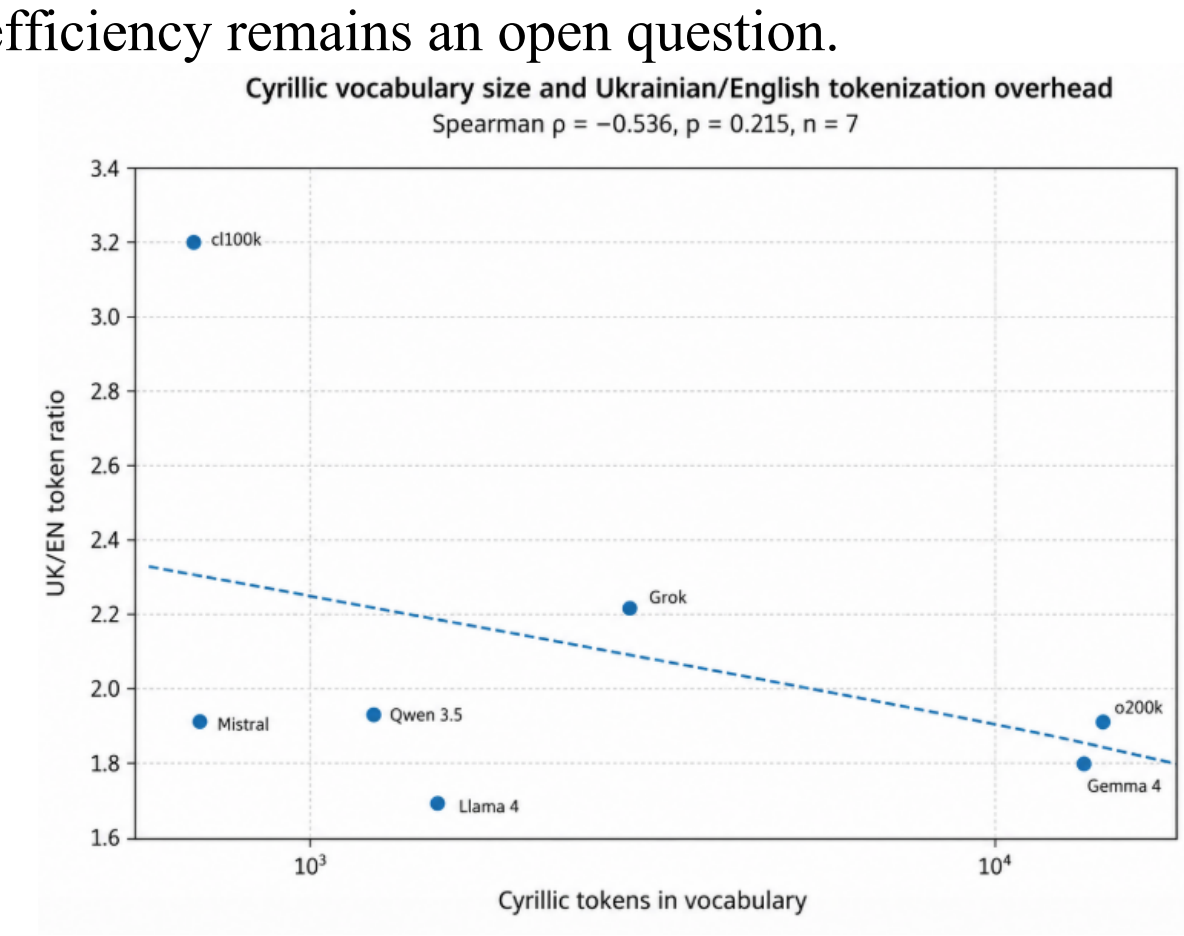

*Figure 1: Association between Cyrillic vocabulary size and tokenization overhead in the complete-case subset of tokenizers with both Cyrillic vocabulary statistics and independently verified UK/EN ratios (Spearman ρ = −0.536, p = 0.215, n = 7). Claude and Qwen 3 are excluded because they do not satisfy the complete-case inclusion criterion for this correlation analysis. Llama 4 remains a notable exception, indicating that raw Cyrillic vocabulary size alone does not fully explain tokenization efficiency.*

## 4.2 Romanization

Table 2 shows the effect of romanizing Ukrainian text using three official transliteration systems. On five of six modern tokenizers, romanization increases token count. The effect is strongest on Llama 4 (+9.5% to +19.0%), the tokenizer with the best native Cyrillic handling, and weakest on Grok (−3.6% to +7.6%), which has the highest UK/EN overhead among the six tokenizers in this comparison.

| System | o200k | Llama 4 | Gemma 4 | Qwen 3.5 | Mistral | Grok |
|---|---|---|---|---|---|---|
| KMU 55 (passport) | +2.4% | +9.5% | +5.8% | +2.4% | +3.8% | −3.6% |
| DSTU A (diacritics) | +10.8% | +19.0% | +15.5% | +10.4% | +12.0% | +7.6% |
| DSTU B (digraphs) | +10.2% | +17.5% | +13.8% | +9.4% | +11.6% | +4.0% |

*Table 2: Romanization impact on 2,787,452 Ukrainian word forms. Positive values indicate more tokens than Cyrillic.*

This result may appear to contradict earlier findings that Latin transliteration reduces token counts for Cyrillic languages. The distinction is between tokenizer generations. Legacy tokenizers (cl100k, GPT-2 era) had minimal Cyrillic vocabulary and benefited from any Latin input. Modern byte-level BPE tokenizers have learned efficient Cyrillic byte merges through exposure to Cyrillic web content. Romanization destroys these learned merges and produces unfamiliar byte sequences, resulting in worse fragmentation.

Table 3 confirms this pattern across five languages. In every case, the tokenizer favors whichever script dominates the web for that language: Cyrillic for Ukrainian and Kazakh (Cyrillic-dominant web), Latin for Serbian and Azerbaijani (Latin-dominant web).

| Language | Words | Script status | Cheaper script | Web dominant |
|---|---|---|---|---|
| Ukrainian | 2,787,452 | Cyrillic only | Cyrillic | Cyrillic |
| Serbian | 2,051,643 | Both active | Latin | Latin |
| Moldovan | 1,555,360 | Latin since 1989 | Latin | Latin |
| Kazakh | 1,173,925 | Mid-transition | Cyrillic | Cyrillic |
| Azerbaijani | 801,603 | Latin since 2001 | Latin | Latin |

*Table 3: Five-language script comparison (o200k tokenizer). The cheaper script matches the dominant web script in all cases.*

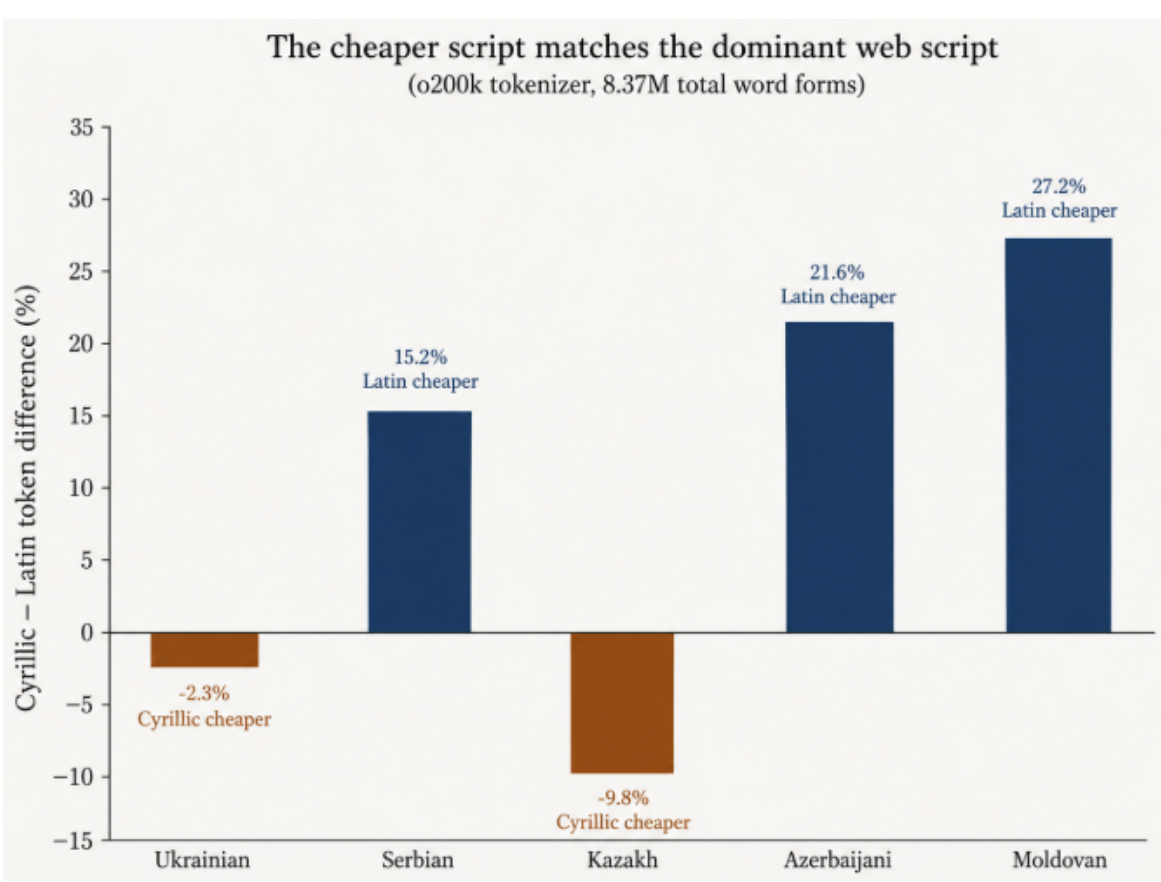

*Figure 2: Cyrillic vs Latin tokenization overhead across five languages (o200k). Values computed as (Cyrillic tokens − Latin tokens) / Latin tokens × 100%. Positive = Cyrillic requires more tokens (Latin cheaper).*

### 4.3 Prompt Compression

LLMLingua-2 reduces Ukrainian e-commerce RAG contexts by 46.8–49.0% across all six tested tokenizers (Table 4). The reduction is broadly consistent with rates reported for LLMLingua-2 on English benchmarks, although differences in datasets and evaluation protocols prevent a direct cross-linguistic comparison.

English token counts were not measured directly because translating the product catalogue could introduce unverifiable changes in text length and vocabulary. Instead, we estimated the English baseline by applying the full-text fertility ratios obtained in Study 1 from the BrUK and Brown corpora. Before compression, the estimated Ukrainian overhead ranges from 68% for Llama 4 to 121% for Grok.

| Tokenizer | UK raw tokens /query | UK compressed tokens /query | Reduction | Estimated EN tokens /query | Raw UK overhead |
|---|---|---|---|---|---|
| Gemma 4 | 1,509 | 793 | 47.4 % | 848 | 78% |
| Grok | 1,787 | 923 | 48.4 % | 810 | 121 % |
| Llama 4 | 1,343 | 689 | 48.7 % | 800 | 68% |
| Mistral Small | 1,631 | 868 | 46.8 % | 855 | 91% |
| o200k | 1,450 | 740 | 49.0 % | 764 | 90% |
| Qwen 3.5 | 1,593 | 816 | 48.8 % | 831 | 92% |

*Table 4: Ukrainian token counts before and after LLMLingua-2 compression, with the estimated English baseline and uncompressed tokenization overhead. English values were estimated using the full-text fertility ratios from Study 1.*

We evaluate compression independently from retrieval. The retriever surfaced the products required to answer 80 of 145 queries (55%); retrieval performance is orthogonal to the tokenization question and is reported only to identify evaluable cases. Among these 80 cases, no compression-induced value losses were observed — all checked prices, ratings, brand names, and availability status were retained. End-to-end validation through Gemini 2.5 Flash confirmed these results: all 80 retrievable queries were sent to the language model with both raw and compressed contexts. The model produced correct answers for 74 queries from raw context and 74 from compressed context — the same 74 in both conditions, with zero compression-induced degradation. The 6 incorrect answers were identical in both conditions, attributable to the language model rather than to compression.

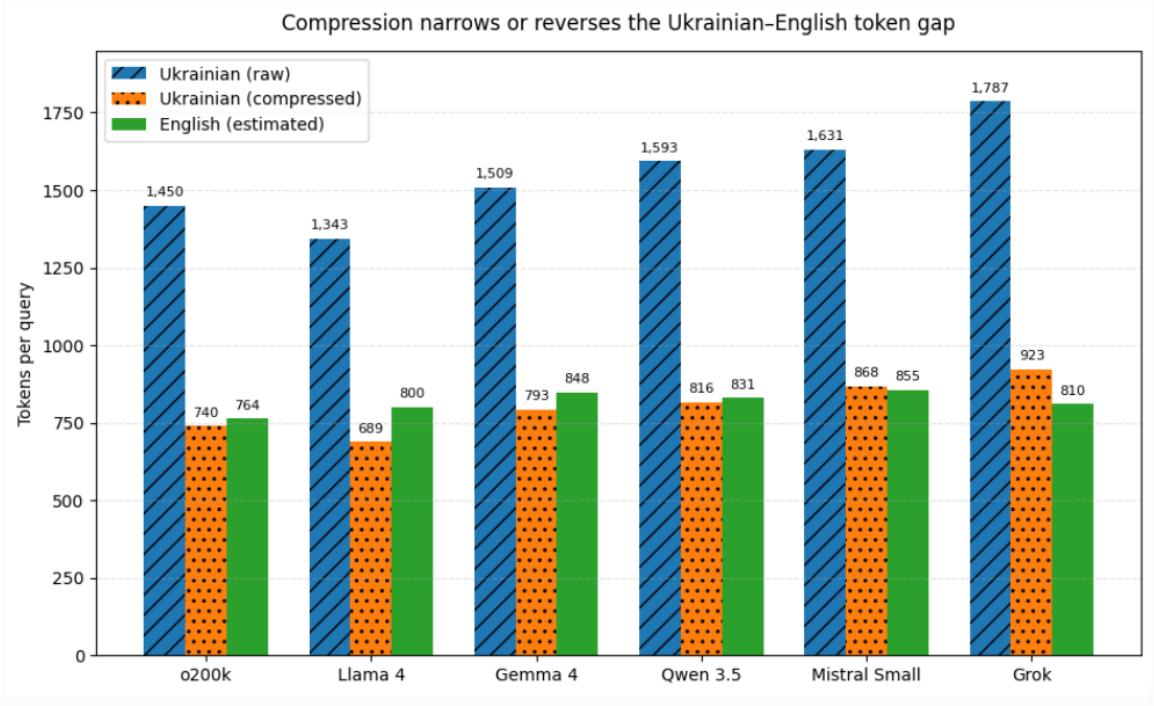

*Figure 3: Ukrainian token counts before and after LLMLingua-2 compression compared with the estimated uncompressed English baseline. Compression brings Ukrainian below the estimated English baseline for o200k, Llama 4, Gemma 4, and Qwen 3.5, while Mistral and Grok remain above it.*

### 4.4 Balanced Tokenizer

Table 5 shows results for balanced BPE tokenizers trained on proportional English–Ukrainian data at four requested vocabulary caps. The production-tokenizer ratios in this section differ from the full-text corpus ratios in Table 1 because Table 5 uses word-by-word encoding on the matched 5K held-out test set, whereas Table 1 reports full-text fertility on BrUK and Brown. At every tested vocabulary size, the balanced tokenizer achieves a lower UK/EN ratio than the selected production tokenizers measured under the same word-by-word held-out protocol. With a 200K vocabulary cap, the balanced tokenizer converges at 158,184 actual entries and achieves 1.30×, compared to 2.22× for production o200k on the same held-out test data. The rounded ratios are non-increasing as the requested vocabulary cap increases (1.34× at 32K, 1.32× at 50K, 1.32× at 100K, and 1.30× under the 200K cap).

| Configuration | UK tok/w | EN tok/w | UK/EN | Vocab |
|---|---|---|---|---|
| Balanced (ours) | 2.049 | 1.582 | 1.30× [1.27, 1.32] | 158,184 actual (200K cap) |
| Balanced (ours) | 2.106 | 1.600 | 1.32× | 100,000 |
| Balanced (ours) | 2.294 | 1.734 | 1.32× | 50,000 |
| Balanced (ours) | 2.431 | 1.813 | 1.34× | 32,000 |
| o200k (production) | 2.784 | 1.256 | 2.22× [2.18, 2.26] | 200,000 |
| Llama 4 (production) | 2.511 | 1.253 | 2.00× | 128,000 |
| Grok (production) | 2.805 | 1.074 | 2.61× | 131,000 |

*Table 5: Balanced vs production tokenizers. All measured with the same protocol: word-by-word encoding on 5,000 held-out test words per language (80/20 split, seed 42). The best balanced tokenizer was trained with a 200K vocabulary cap and converged at 158,184 actual entries. 95% bootstrap CIs are shown for the key comparison between this 200K-cap balanced tokenizer and o200k (10,000 resamples, $p < 0.0001$ paired bootstrap).*

The improvement cannot be explained by a larger vocabulary: the best balanced model was trained with a 200K cap and converged at fewer actual entries than production o200k, yet it still achieved a substantially lower UK/EN ratio. It comes from vocabulary allocation. Production tokenizers trained on English-dominated web crawls allocate a small fraction of merge rules to Ukrainian byte sequences. Balanced training allocates vocabulary proportionally, giving Ukrainian the coverage it needs. The 1.30× ratio under the 200K vocabulary cap is an empirical result of balanced training, not a theoretical lower bound. The 100% Ukrainian configuration in Table 6 reaches a ratio below 1.0, confirming that UTF-8 byte length does not impose a fixed tokenization floor. The overhead is determined by training data distribution and learned merge rules, not by the physical number of bytes per character.

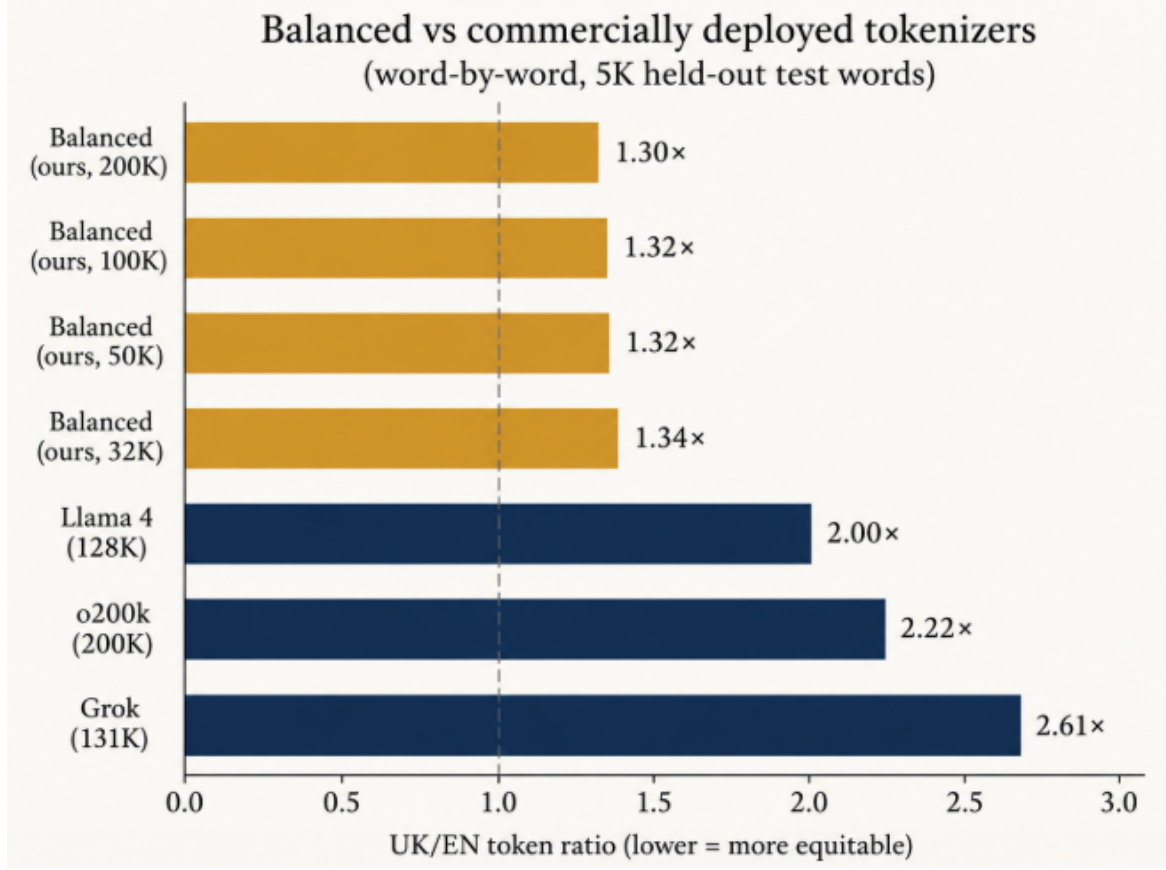

*Figure 4: UK/EN ratio for balanced tokenizers trained in this study and three selected production tokenizers (o200k, Llama 4, Grok). The best balanced model used a 200K vocabulary cap and converged at 158,184 actual entries. All values were measured with word-by-word encoding on the same 5K held-out test set.*

#### 4.4.1 *Training Data Balance Sweep*

To characterize the trade-off across different Ukrainian shares of training data, we trained tokenizers at 11 ratios from 0% to 100% Ukrainian. Table 6 shows the full tradeoff. At 0% Ukrainian share, the tokenizer has no Cyrillic merge rules and encodes each Ukrainian character as two separate byte tokens, producing 11.89 tokens per word (7.89× ratio). Adding just 10% Ukrainian data collapses this to 1.55× — already 30% lower than production o200k (2.22×). At 50%, the ratio reaches 1.32× at 100K vocabulary (41% below production). Beyond 80%, English degradation accelerates: at 100%, English tok/word rises by 55% (2.340 vs 1.507 at 0%) while Ukrainian gains diminish.

| UK share | UK tok/w | EN tok/w | UK/EN | vs production (2.22×) |
|---|---|---|---|---|
| 0% | 11.891 | 1.507 | 7.89× | 255% higher |
| 10% | 2.371 | 1.531 | 1.55× | 30% lower |
| 20% | 2.234 | 1.549 | 1.44× | 35% lower |
| 30% | 2.166 | 1.560 | 1.39× | 37% lower |
| 40% | 2.140 | 1.57 | 1.36× | 39% |

| | | 1 | | lower |
|---|---|---|---|---|
| 50% | 2.106 | 1.600 | 1.32× | 41% lower |
| 60% | 2.092 | 1.628 | 1.29× | 42% lower |
| 70% | 2.060 | 1.666 | 1.24× | 44% lower |
| 80% | 2.043 | 1.695 | 1.21× | 45% lower |
| 90% | 2.023 | 1.755 | 1.15× | 48% lower |
| 100% | 1.995 | 2.340 | 0.85× | — |
| Production o200k | 2.784 | 1.256 | 2.22× | baseline |

*Table 6: Balance sweep at 100K vocabulary (word-by-word encoding, 5K test words per language, full 256-byte alphabet). At 0%, each Cyrillic character produces two byte tokens. Adding 10% Ukrainian data introduces basic merges, reducing the ratio by 80%.*

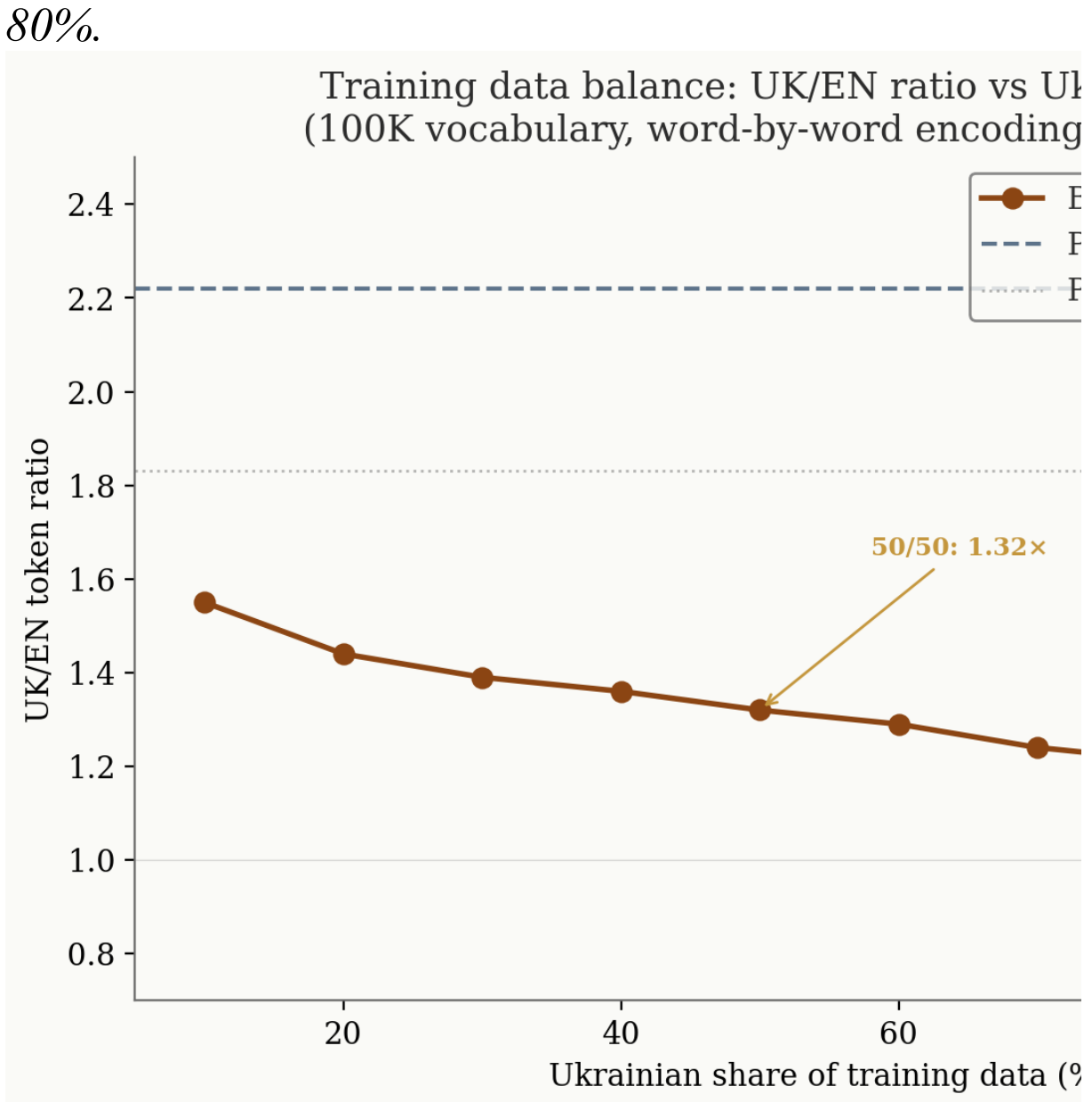

*Figure 5: Training data balance sweep (100K vocabulary, word-by-word encoding). UK/EN ratio decreases monotonically with more Ukrainian data. At 100%, the ratio falls below 1.0 (0.85×), showing that UTF-8 byte length does not impose a UK/EN ratio above 1.0 under this experimental setup. Production o200k (2.22×) and Llama 4 (2.00×) shown for reference. The 0% condition (7.89×) is omitted from the plot to preserve the scale of the 10–100% range.*

### 4.4.2 Fertility Distribution Shift

The balanced tokenizer does not merely reduce the mean: it shifts the entire fertility distribution. On 5,000 Ukrainian test words, the share of single-token words increases from 22.7% (production o200k) to 34.1% (balanced 100K). Words requiring 4 or more tokens drop from 30.3% to 9.4%. Of 5,000 test words, 2,524 improved (fewer tokens), only 77 degraded (more tokens), and 2,399 were unchanged — a 33:1 improvement-to-degradation ratio.

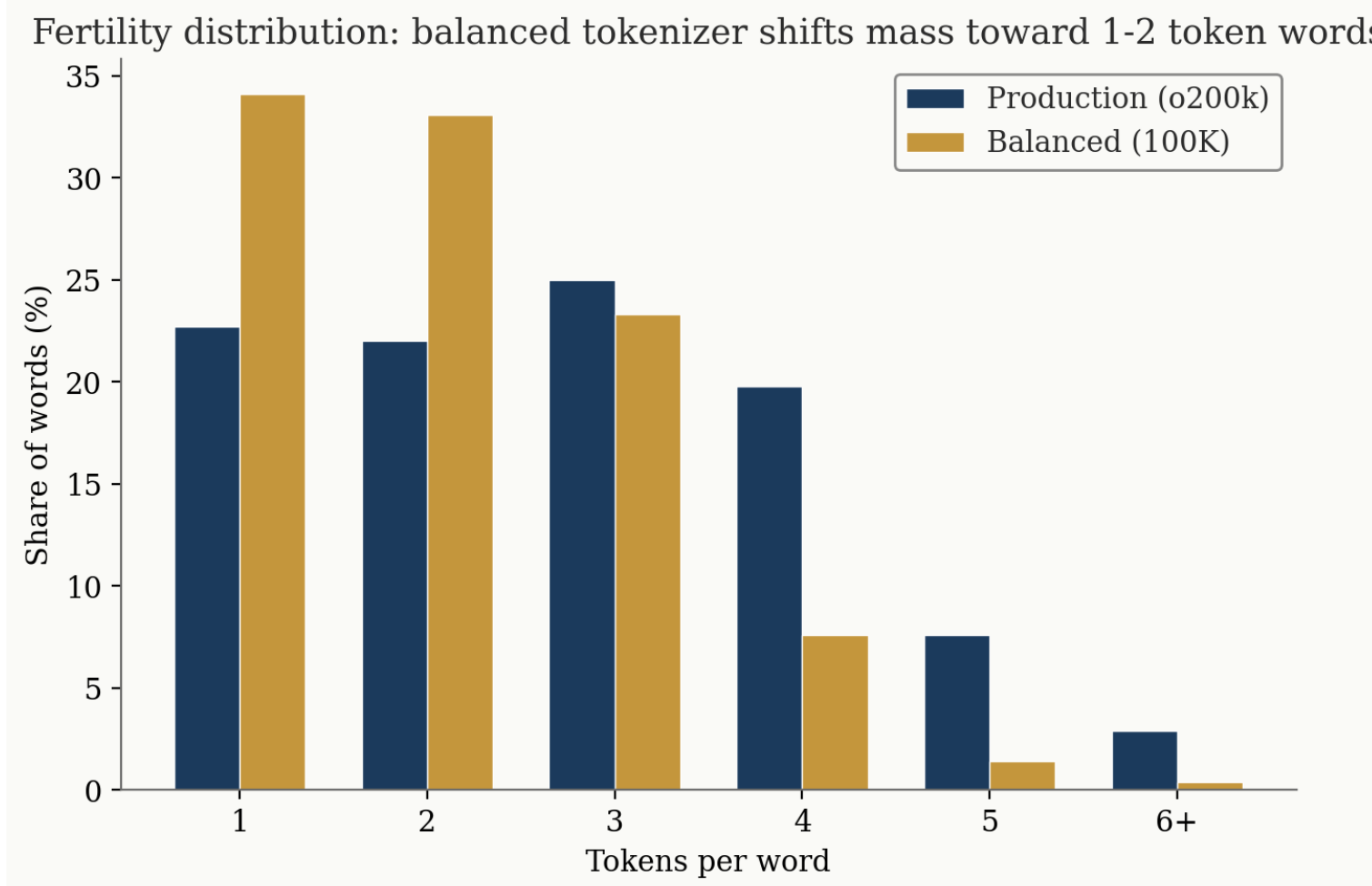

*Figure 6: Fertility distribution for 5,000 Ukrainian test words. The balanced 100K tokenizer shifts probability mass from words requiring four or more tokens toward one- and two-token words.*

### 4.4.3 Merge Candidate Analysis

A Rust-based corpus profiler identified the specific Ukrainian word forms with the highest tokenization cost (Appendix B). Adding the top 100 words as single tokens to the Qwen 3 tokenizer would reduce total tokenization by an estimated 7.6%; the top 500 by 18.1%. These are tokenizer-level upper bounds calculated as freq × (tokens − 1); integrating new tokens into an existing model would require embedding adaptation or full retraining. The most expensive words are common function words and inflected forms (після, україни, його, також, було) — high-frequency items that appear in nearly every Ukrainian text.

## 5 Discussion

Our results support three conclusions. First, multiple lines of evidence suggest that training data composition is an important contributor to Cyrillic tokenization overhead, rather than the script's UTF-8 byte length alone. The cross-linguistic evidence (Table 3) is consistent with the interpretation that tokenizers favor the script most prevalent in available web data for a given language, while the balanced tokenizer experiment (Table 5) demonstrates that proportional training data allocation can reduce the overhead from 2.22× to 1.30× under a 200K vocabulary cap, with no architectural changes.

Second, prompt compression provides an effective inference-time mitigation. Ukrainian contexts were reduced by 47–49%, while the 80 retrievable cases showed no compression-induced loss in checked values or end-to-end accuracy. The similarity to rates reported on English benchmarks is suggestive, but differences in data and evaluation prevent a direct cross-linguistic entropy comparison.

Third, for languages where modern tokenizers have already learned Cyrillic byte merges (such as Ukrainian), romanization is counterproductive. More broadly, across all five tested languages, tokenization efficiency favors whichever script dominates the web — a pattern that has practical implications for the ongoing script reform debate in Kazakhstan and the periodic Latinization proposals for Ukrainian. Under current tokenizer distributions, switching scripts could increase AI costs in the short to medium term, at least until sufficient Latin-script web content accumulates and future tokenizers are retrained on it.

A limitation of the causal interpretation is that the correlation between Cyrillic vocabulary size and tokenization overhead ($\rho = -0.536$, $p = 0.215$) is based on seven tokenizers with complete UK/EN ratio data. The association is suggestive but does not reach conventional significance. The within-lab generational improvements (e.g., OpenAI cl100k → o200k: −41% with vocabulary expansion from 729 to 14,208 Cyrillic tokens) and the balanced tokenizer experiment provide complementary evidence, but a controlled study with more tokenizers would strengthen the claim.

## 6 Conclusion

We have shown that tokenization overhead for Ukrainian is substantial, reaching 68–121% on modern tokenizers and up to 220% on the older cl100k tokenizer. Two mitigation strategies proved effective: prompt compression reduced Ukrainian input length by 47–49% without observed losses among retrievable cases, while balanced vocabulary allocation reduced the UK/EN ratio from 2.22× to 1.30× under a 200K vocabulary cap, with the tokenizer converging at 158,184 actual entries (paired bootstrap, $p < 0.0001$). The results suggest that training data allocation is an important contributor to tokenization overhead and that UTF-8 byte length alone does not determine tokenization efficiency. We release the benchmark dataset, the balanced tokenizer, and the corpus profiler to support further research on equitable tokenization for underrepresented scripts.

## Data and Code Availability

For review, we provide an anonymized supplementary archive containing the processed benchmark tables, tokenizer configurations, trained balanced-tokenizer files, corpus-profiling outputs, and scripts required to reproduce the reported results.

## 7. Limitations

English cost estimates use empirical ratios from corpus comparison, not parallel translations. This is a deliberate choice to avoid translation artifacts, but it means English figures were estimated by applying corpus-level fertility ratios to the RAG-specific Ukrainian token counts. The e-commerce benchmark represents one domain; other domains (legal, medical) may show different Cyrillic/Latin content ratios. The balanced tokenizer is a proof-of-concept: we did not evaluate downstream LLM performance with the modified vocabulary. Compression was evaluated on 80 paired end-to-end query cases, comprising 160 generated responses: 80 from raw contexts and 80 from compressed contexts. A full subjective quality evaluation across compression rates is a separate study. Only one compression method (LLMLingua-2) was tested; alternative approaches may yield different results on Cyrillic text.

## Ethical Considerations

This study uses public or research-oriented corpus data, tokenizer outputs, and structured product information. No personal data, private messages, user profiles, or sensitive individual-level information were collected or analysed. The study aims to support more equitable AI access for underrepresented-script language communities.

## Appendix A: E-Commerce Benchmark Details

| **Category** | **Products** | **Avg doc length (chars)** |
|---|---|---|
| headphones | 264 | 737 |
| laptops | 158 | 761 |
| monitors | 124 | 931 |
| refrigerators | 120 | 786 |
| routers | 126 | 750 |
| smartphones | 224 | 541 |
| tablets | 106 | 574 |
| tvs | 140 | 785 |
| vacuum_cleaners | 154 | 722 |
| washing_machines | 120 | 852 |

*Table A1: Knowledge base composition (1,536 products, 10 categories).*

The knowledge base contains 65.7% Cyrillic characters, 10.4% Latin characters (brand names, model numbers), 5.0% digits (prices, specifications), and 18.9% other (punctuation, whitespace). Latin characters and digits constitute 15.4% of the knowledge-base characters. Although these characters are not Cyrillic, their tokenization may still depend on the surrounding textual context.

## Appendix B: Merge Candidate Analysis

| **Word** | **Tokens** | **Freq** | **Current cost** | **Savings if 1 token** |
|---|---|---|---|---|
| країни | 4 | 1,820 | 7,280 | 5,460 |
| його | 2 | 3,487 | 6,974 | 3,487 |

| вони | 3 | 2,105 | 6,315 | 4,210 |
|---|---|---|---|---|
| якщо | 4 | 1,523 | 6,092 | 4,569 |
| навіть | 4 | 1,397 | 5,588 | 4,191 |
| також | 3 | 1,709 | 5,127 | 3,418 |
| тільки | 4 | 1,245 | 4,980 | 3,735 |
| було | 2 | 2,405 | 4,810 | 2,405 |
| після | 4 | 1,177 | 4,708 | 3,531 |
| вона | 2 | 2,322 | 4,644 | 2,322 |

*Table B1: Ten most expensive Ukrainian words (Qwen 3, Rust profiler on BrUK). Current cost = freq × tokens. Savings = freq × (tokens − 1). Top 100 words as single tokens: −7.6% of total tokenization. Top 500 (out of 24,719 unique words with frequency ≥ 5 in the full BrUK corpus): −18.1%. These are tokenizer-level upper-bound estimates; adding tokens to an existing model would require embedding adaptation or retraining.*